# Generative Chemical Transformer: Neural Machine Learning of Molecular Geometric Structures from Chemical Language via Attention

Hyunseung Kim[†], Jonggeol Na[‡,*], Won Bo Lee[†,*]

[†]School of Chemical and Biological Engineering, Seoul National University, Gwanak-ro 1, Gwanak-gu, Seoul 08826, Republic of Korea

[‡]Department of Chemical Engineering and Materials Science, Graduate Program in System Health Science and Engineering, Ewha Womans University, Seoul 03760, Republic of Korea

**Correspondence and requests for materials should be addressed to**

J.N. (email: jgna@ewha.ac.kr) or W.B.L. (email: wblee@snu.ac.kr).

## ABSTRACT:

Discovering new materials better suited to specific purposes is an important issue in improving the quality of human life. Here, a neural network that creates molecules that meet some desired conditions based on a deep understanding of chemical language is proposed (Generative Chemical Transformer, GCT). The attention mechanism in GCT allows a deeper understanding of molecular structures beyond the limitations of chemical language itself which cause semantic discontinuity by paying attention to characters sparsely. It is investigated that the significance of language models for inverse molecular design problems by quantitatively evaluating the quality of the generated molecules. GCT generates highly realistic chemical strings that satisfy both chemical and linguistic grammar rules. Molecules parsed from generated strings simultaneously satisfy the multiple target properties and vary for a single condition set. These advances will contribute to improving the quality of human life by accelerating the process of desired material discovery.

# ■ INTRODUCTION

Material discovery is a research field that searches for new materials that fit specific purposes. Searching molecular structures that simultaneously satisfy the multiple desired target conditions requires a high level of expert knowledge and considerable time and cost, since the chemical space is vast; the number of organic molecules less than 500 Da exceeds $10^{60}$.[1] For this reason, inferring the desired molecules using artificial intelligence with transcendental learning can help to accelerate the process of material discovery.

In recent years, attempts have been made to utilize artificial neural networks in chemistry fields where the relationship between molecular structure and physical properties is complex. To learn the molecular structures in neural nets, the molecular structure must be expressed as structured data. Molecules are made up of atoms and connecting bonds, so they are similar to graphs. However, techniques for expressing molecular structures in the form of strings also have been studied to facilitate the construction and utilization of molecular information as a database, since the form of string data is more convenient to handle than the form of graph data.[2-6] The Simplified Molecular-Input Line-Entry System (SMILES)[4-6] developed in the 1980s is the most popular artificial language used to express molecular structures in detail. Similar to natural language, SMILES also has an arrangement of characters according to grammar and context.

Several approaches have demonstrated that Natural Language Processing (NLP) models are applicable to inverse molecular design problems by generating molecules expressed in chemical language via a language model to select a character that follows the currently generated string,[7-11] a language model combined with a Variational Autoencoder (VAE)[12] to compress molecular information into latent space and re-sample the latent code to create various strings,[13,14] a language model combined with a Generative Adversarial Network (GAN)[15] to create strings from noise,[16] or a language model combined with reinforcement learning to reward a natural character that follows the currently generated string.[17] However, why language model-based molecular generators work well has not been comprehensively analyzed.

An important point in material discovery is to search the molecular structures that meet multiple desired target conditions. The desired molecules can be discovered in 2 steps by applying additional optimization or navigation process to the generative model: Bayesian optimization,[13,18] particle swarm optimization,[19] genetic optimization,[20-22] or Monte Carlo tree search.[23-25] Another method is to use conditional models which create molecules with the given target conditions in 1 step. The latter can shorten the time consumed to discover the desired molecules and directly control the molecules to be generated. For this reason, attempts have been made to directly input conditions to a recurrent neural network (Conditional Recurrent Neural Network, cRNN[26]). Unlike generative models, since a Recurrent Neural Network (RNN) is based on one-to-one matching of input and output, there is a limitation in using it to infer the various molecular candidates with a single condition set; here, the generative models refer to models that can output various results by decoding latent codes sampled from the distribution of latent variables (latent space).

Here, Generative Chemical Transformer (GCT), which embeds Transformer[27]—a state-of-the-art architecture that became a breakthrough for NLP problems by using an attention mechanism[28]—into a conditional variational generative model is proposed. From the point of view of data recognition and processing, GCT is close to a conditional variational generator that embodies the language recognition ability of the attention mechanism in Transformer. To use GCT for material discovery, it is intended to take advantage of both the high-performance language model and the conditional generative model. GCT is analyzed by quantitatively evaluating the generated molecules and investigate the significance of language models for inverse molecular design problems. It is shown that a deep understanding of the molecular geometric structure learned from chemical language by paying attention to each character in chemical strings sparsely helps to generate chemical strings satisfying the chemical valance rule and syntax of the chemical language (grammar understanding). The strings are parsed into highly realistic molecules (context understanding). Additionally, it is demonstrated that the conditional variational generator, which is the skeleton of GCT, helps to generate molecules that satisfy multiple preconditions simultaneously (conditional generator) and varies for a single precondition set (variational generator). In addition, the autoencoder, a substructure of GCT, makes the molecular size controllable.

## ■ MATERIALS AND METHODS

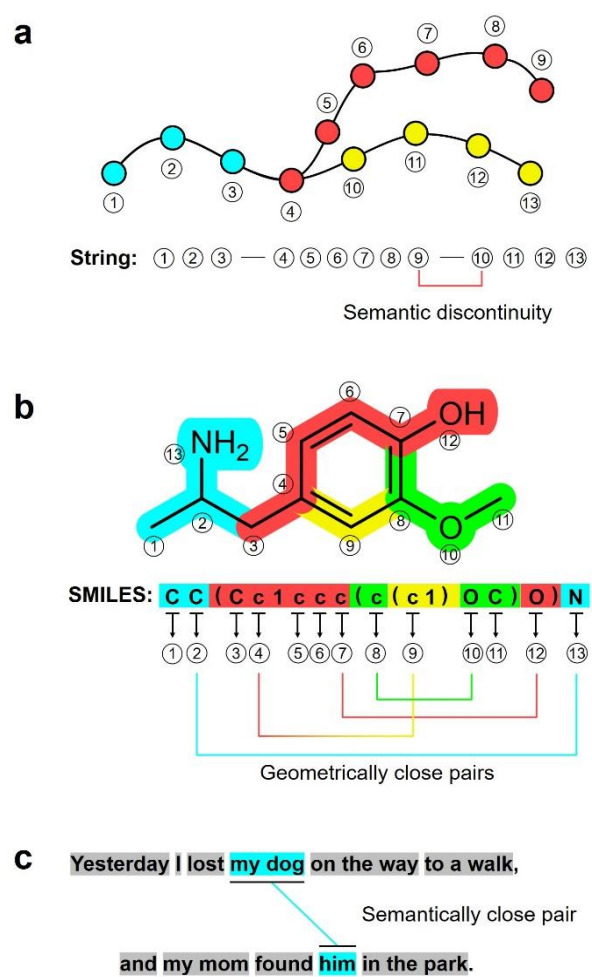

**Figure 1.** Structural limitation of language forms. (a) An example of a non-Hamiltonian graph. A Hamiltonian graph has a path that passes through all the points in the graph only once, and a non-Hamiltonian graph does not have such a path. (b) An example of a non-Hamiltonian molecular graph and its SMILES string: 4-(2-aminopropyl)-2-methoxyphenol. Each atom is labelled with a circled number. Different colors refer to different branches. (c) In natural language, words that are semantically close within a sentence are not always structurally

close within a sentence.

**Overcoming Structural Limits of Language via Attention.**

It is expected that introducing an attention mechanism to language-based inverse molecular design problems can help networks understand the geometric structures of molecules beyond the limitations of chemical language itself. SMILES, a chemical language, represents molecules as one-dimensional text. It is a powerful language since a one-SMILES string is converted into one exact molecule. Unfortunately, however, since most molecules are non-Hamiltonian graphs, it is self-evident that semantic discontinuity occurs whenever a branch in a molecule is translated into a one-dimensional string ([Figure 1](#)a). SMILES distinguishes each branch with open and close parentheses and creates a gap between the distance of two characters within the string and the distance of the corresponding two atoms in the molecular graph. An example is shown in [Figure 1](#)b. Even though atom2 and atom13 in [Figure 1](#)b are geometrically adjacent, they are far from each other in the string. In other words, the longer the branch is, the harder it is to imagine the molecular structure by reading the chemical string in order (similar to memory cells). A similar phenomenon occurs in natural language ([Figure 1](#)c). The longer the sentence is, the larger the gap between the words within the sentence and the semantic similarity. Unfortunately, in chemical language, there are more areas where semantic discontinuity occurs. In the field of NLP, by introducing an attention mechanism to this problem, language models are able to pay sparse attention to semantically related parts; it is a departure from the traditional way of perceiving context in the order of sentences (memory cells). Transformer is an architecture involving an attention mechanism in the form of a neural network, and it became a breakthrough in cognitive ability. This study was started with the expectation that the attention mechanism could help structural understanding beyond the semantic discontinuity of chemical language. The results and discussion for this problem will be covered in a later section.

**SMILES Language.**

SMILES expresses the atoms, bonds, and rings that make up molecules as a string. An atom is represented by the alphabet of the element symbol, and a bond is represented by a single bond (-), double bonds (=), and triple bonds (#), depending on the type. In general, a bond that can be easily inferred through the atoms or ring structure of the surrounding atom is omitted. The notation of hydrogen is also omitted in SMILES string if single bond hydrogen can be explicitly inferred by the chemical valence rule, however, single bond hydrogen can be clearly indicated by using [H] if the bond is implicit. For charged atoms, where the number of hydrogen bonds cannot be determined explicitly, atoms and formal charges are written together in brackets [ ]. The beginning and end of each ring are expressed with the same digit, and the pair must be correct; if a ring is open, it must be closed. The atoms present in the aromatic ring are written in lowercase, while the atoms outside the ring are capitalized. The branches in molecules are indicated by opening and closing parentheses (see **Figure 1b**). A more detailed description of SMILES is in the ref. [4-6].

**Tokenization and Token Embedding.**

To input SMILES string to language models, the process of tokenizing by semantic units is necessary. The SmilesPE[29] tokenizer was used to tokenize the SMILES strings included in the training data of Molecular Sets benchmarking platforms (MOSES).[30] In total, 28 types of tokens are used: 4 special tokens (<unknown>, <pad>, <sos>, and <eos>), 13 atom tokens (<C>, <c>, <O>, <o>, <N>, <n>, <F>, <S>, <s>, <Cl>, <Br>, <[nH]>, and <[H]>), 3 bond tokens (<->, <=>, and <#>), 2 branch tokens (<(> and <)>), and 6 ring tokens (<1>, <2>, <3>, <4>, <5>, and <6>). Note that tokens related to charged atom (e.g. <[O-]>, <[n+]>) and tokens related to stereochemistry (e.g. </> , <\>) were not considered as they are not covered by MOSES database. Each token that constitutes the SMILES string is one-hot encoded in 28 dimensions and embedded in 512 dimensions. The condition of GCT is also embedded in 512 dimensions.

**Attention Mechanism.**

The attention mechanism is the core of Transformer's language cognition abilities. The attention mechanism allows Transformer to self-learn which token of the input string is better to focus on to perform a given task better. The attention mechanism uses three vectors: the query $Q$, the key $K$, and the value $V$. The attention mechanism calculates the similarity between Q and all keys in K, and the calculated similarity is multiplied by the value corresponding to the key to calculate the attention scores. The scale-dot attention used in the Transformer is the same as eq 1 [27]:

$$\text{Attention}(Q, K, V) = \text{Softmax}\left(\frac{QK^T}{\sqrt{d_k}}\right)V \quad (1)$$

where $d_k$ is the dimension of $K$ and $d_k$ must correspond to the dimension $d_q$ of $Q$.

The Transformer uses multi-head attention instead of single-head attention (eq 2)[27]:

$$\text{MultiHead}(Q, K, V) = \text{Concat}(head_1, \dots, head_h)W^O \quad (2)$$

$$where\ head_i = \text{Attention}(QW_i^Q, KW_i^K, VW_i^V)$$

Where $W_i^Q \in \mathbb{R}^{d_{model} \times d_k}$, $W_i^K \in \mathbb{R}^{d_{model} \times d_k}$, $W_i^V \in \mathbb{R}^{d_{model} \times d_k}$, $W_i^O \in \mathbb{R}^{hd_v \times d_{model}}$ and $d_{model} = d_k \times h = d_v \times h$. Here, $h$ is the number of multi-head and $d_v$ is the dimension of $V$. Each head ($head_i$) calculates an attention score between $Q$ and $K$ from different viewpoints using the different weights belonging to each head ($W_i^Q$, $W_i^K$, and $W_i^V$).

**Generative Chemical Transformer.**

We designed GCT, an architecture that embeds Transformer—one of the most advanced language models—into a Conditional Variational Autoencoder (cVAE),[31] which creates molecules with target properties based on a deep understanding of chemical language. Transformer, the core of GCT's language recognition ability, is mainly used as a Neural Machine Translator (NMT). It consists of an encoder and a decoder (**Figure 2**a). The encoder receives a sentence to be translated and understands the received sentence through self-attention. Then, the processed information from sentence comprehension is passed to the decoder. The decoder iteratively selects the next token that will follow the translated sentence up to this point, referring to the information received from the encoder and the sentences translated up to the previous step; if there is no translated sentence at the beginning of translation, the special token 'start of sentence <sos>' is used. The decoder uses the input information to iteratively select the next token that will follow the translated sentence up to the previous step. Finally, the translation ends when the decoder selects a special token 'end of sentence <eos>'.

GCT is a structure that inserts a low-dimensional conditional Gaussian latent space between the encoder and the decoder of the Pre-Layer Normalization (Pre-LN) Transformer.[32] (**Figure 2**b). Pre-LN Transformer is a modified version of the original (Post-Layer Normalization, Post-LN) Transformer. The combination of language models and variational autoencoders is vulnerable to posterior collapse.[33] A complete solution to posterior collapse has yet to be identified; however, it is known that Kullback-Leibler divergence (KL) annealing can alleviate this problem.[34] Since KL annealing (KLA) controls the gradient size, adopting Pre-LN Transformer—designed to stabilize the gradient flow of the (Post-LN) Transformer—can facilitate KLA manipulation. The loss function of GCT is as follows:

$$L(\emptyset, \theta; x_{enc}, x_{<t}, x_t, c) = k_w D_{KL}(q_\emptyset(z|x_{enc}, c) \parallel p(z|c)) - E_{q_\emptyset(z|x_{enc}, c)}[\log g_\theta(x_t|z, x_{<t}, c)] \qquad (3)$$

where $D_{KL}(\cdot)$ is the KL divergence, and $E[\cdot]$ is the expectation of $\cdot$. $q_\emptyset$ is a parameterized encoder function, $g_\emptyset$ is a parameterized decoder function (generator), $p(\cdot|c)$ is a conditional Gaussian prior. Here, $\emptyset$, $\theta$, $x_{enc}$, $x_{<t}$, $z$, $x_t$, $c$, $k_w$ are the parameter set of the encoder, the parameter set of the decoder, the input of the encoder, the input of the decoder, the latent variables, the reconstruction target, the conditions, and the weight for KLA, respectively. The encoder and decoder each consist of six Pre-LN Transformer blocks. Each block has dimensions of 512 and 8-head attention, and the dimension of the feed-forward block is 2,048. The Gaussian latent space is designed in 128 dimensions.

The self-attention block of the encoder obtains the concatenated array of the SMILES string and three different properties: the octanol-water partition coefficient (logP), the topological Polar Surface Area (tPSA), and the Quantitative Estimate of Drug-likeness (QED).[35] The encoder-decoder attention block in the decoder obtains the concatenated array of latent code and condition (three properties), and the self-attention block in the decoder obtains only the SMILES string. In the training phase, GCT performs the task of reconstructing the SMILES string—input through the encoder—by referring to the given hints on the molecular properties. In this process, the low-dimensional latent space acts as the model's bottleneck to find as much meaningful information that can be restored to the decoder as possible by exploiting the limited information passed through the bottleneck. Then, meaningful latent variables for molecular structures and properties are represented in the low-dimensional continuous latent space. In the inference phase, a sampled latent code and target properties are input into the learned decoder, and the decoder selects the next tokens iteratively through a 4-beam search; which is a kind of tree search method.

A dropout rate of GCT is 0.3 applied. Learning is conducted by the Adam optimizer.[36] The initial learning rate is $10^{-4}$. The expansion rate of the momentum is 0.9 and the expansion rate of the adaptive term is 0.98. Two methods are applied to schedule the learning rate (GCT-WarmUP and GCT-SGDR in Table 1). One is to use the warm-up scheduler (eq 4) [27]:

$$\eta = 3\,d_{model}^{-0.5} \cdot \min(s_{current}^{-0.5}, s_{warmup}^{-1.5}) \quad (4)$$

where, $\eta$ *means learning rate*, $s_{current}$ means current training step, and $s_{warmup}$ means warm-up steps. $s_{warmup}$ is set to 100,000. The other is to use Stochastic Gradient Descent with warm Restart (SGDR) of one epoch cycle (eq 5) [37]:

$$\eta = \eta_{min} + 0.5(\eta_{max} - \eta_{min})\left(1 + \cos\left(\frac{s_{current}}{s_{cycle}}\pi\right)\right) \quad (5)$$

where, $\eta_{min}$ means minimum learning rate, $\eta_{max}$ means maximum learning rate, $s_{cycle}$ is learning rate scheduling step cycle. Here, $\eta_{min}$, $\eta_{max}$, and $s_{cycle}$ are set to 0, 0.0001, and 1-epoch steps, respectively. KL annealing was applied to increase $k_w$ from 0.02 to 0.50 at 0.02 intervals per epoch for a total of 25 epochs. Other details of GCT are provided in SI.

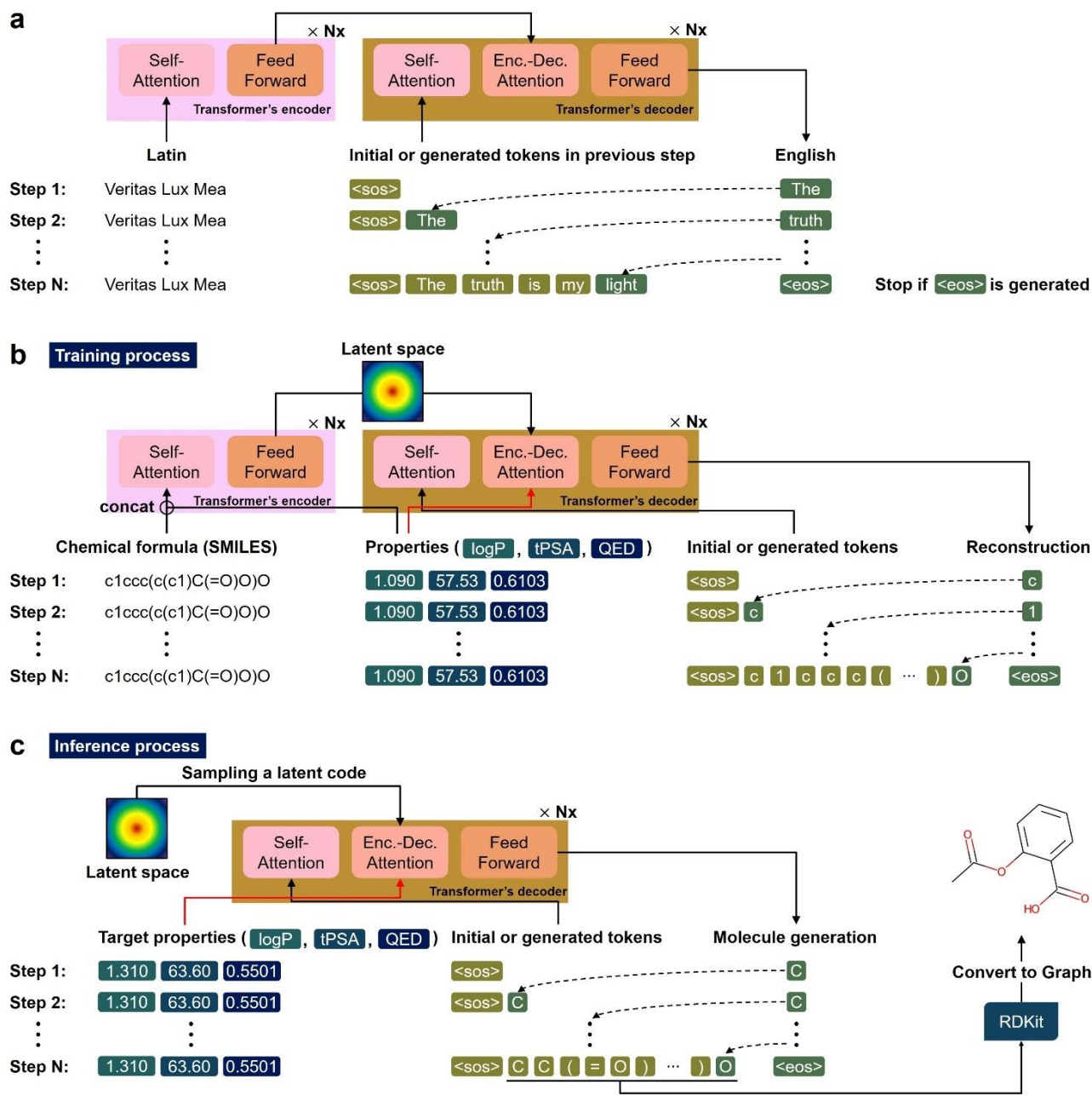

**Figure 2.** De novo molecular generation via GCT. (a) Transformer model for NMT: an example of translating Latin into English. It iteratively selects the next English word by referring to the Latin sentence and the English sentence translated up to the previous step. (b) In the process of learning to reproduce the input chemical formula, GCT learns the molecular structure and three different properties: logP, tPSA, and QED. It represents the information of molecular structure and properties in the latent space during the learning process. (c) The trained GCT generates a de novo molecule that satisfies the target properties by decoding the molecular information sampled from the latent space and the given preconditions.

**Datasets & Benchmark.**

The GCT model was trained and benchmarked using a database of MOSES benchmarking platforms. The MOSES database is a benchmarking dataset for drug discovery created by sampling molecules from the ZINC is Not Commercial (ZINC) database[38]—composed of commercially available compounds—that satisfy specific conditions: a weight in the range from 250 to 350 daltons, number of rotatable bonds is not greater than 7, not containing charged atoms or atoms other than C, N, S, O, F, Cl, Br, and H or cycles longer than 8 atoms. The MOSES database consists of training samples (1.7 M), test samples (176 k), and scaffold test samples (176 k), which have scaffolds that never appear in the training samples. It is also designed to closely match the distribution between the datasets. The three additional properties (logP, tPSA, and QED) computed from RDKit[39] are used for GCT learning.

In general, the quality of network training can be evaluated by measuring how different the model's predicted and the actual labels are. However, for molecular generative models, the small mean loss does not guarantee that the generative model performs well because the artifacts in the generated molecules, which are not observed in the mean loss measurement, may not fit the chemical valence rule or may make the molecules unrealistic. For this reason, the quality of the generated molecules needs to be checked against the following criteria:

- How plausible are the generated molecules?
- Do the generated molecules satisfy the target properties?
- Can multiple candidates be generated for a single precondition set?
- Can de novo molecules be created in a short time?

In total, 30,000 SMILES strings are generated by the trained GCT model and evaluated by MOSES benchmarking score metrics (Table 1). In addition to relatively simple scores such as the validity, uniqueness, internal diversity, filters, and novelty, the MOSES benchmarking platform also provides metrics that can measure similarity with reference molecules such as the Similarity to a Nearest

Neighbor (SNN),[30] Fréchet ChemNet Distance (FCD),[40] Fragment similarity (Frag),[30] and Scaffold similarity (Scaf).[30]

The SNN score is calculated via eq 6:

$$\text{SNN}(G, R) = \frac{1}{|G|} \sum_{m_G \in G} \max_{m_R \in R} T(m_G, m_R) \quad (6)$$

where $G$ and $R$ refer to the set of molecules generated and reference molecules, respectively, $m$ stands for Morgan fingerprints,[41] and $T(A, B)$ stands for the Tanimoto similarity[42] between set $A$ and set $B$.

The FCD uses activation of the penultimate layer in ChemNet and is designed to predict bioactivity. It calculates the difference in the distributions between $G$ and $R$ via eq 7:

$$\text{FCD}(G, R) = \|\mu_G - \mu_R\|^2 + Tr\left(\sum_G + \sum_R - 2\left(\sum_G \sum_R\right)^{1/2}\right) \quad (7)$$

where $\mu$ is the mean, $\Sigma$ is the covariance, and $Tr(\cdot)$ is the trace operator.

The Frag score is calculated via eq 8:

$$\text{Frag}(G, R) = \frac{\sum_{f \in F}(c_f(G) \cdot c_f(R))}{\sqrt{\sum_{f \in F} c_f^2(G)}\sqrt{\sum_{f \in F} c_f^2(R)}} \quad (8)$$

where $F$ is the set of 58,315 unique BRICS fragments,[43] and $C_f(A)$ is the frequency with which fragment $f \in F$ appears in the molecules in set $A$.

The Scaf score is calculated via eq 9:

$$\text{Scaf}(G, R) = \frac{\sum_{s \in S}(c_s(G) \cdot c_s(R))}{\sqrt{\sum_{s \in S} c_s^2(G)}\sqrt{\sum_{s \in S} c_s^2(R)}} \quad (9)$$

where $S$ is the set of 448,854 unique Bemis-Murcko scaffolds[44] and $C_s(A)$ is the frequency at which scaffold $s \in S$ appears in the molecules in set A.

**Condition Sampling.**

The properties considered in this problem are logP, tPSA, and QED. A three-dimensional histogram was derived after dividing each property into 1,000 equal sections between the maximum and minimum values in the training data. Then, the cells were sampled according to the probability that data samples exist in each cell; here, the probability is the number of samples in that cell out of the total samples. Next, uniform noise was added at the center value of the cell to create condition sets for the 30,000 molecules to be generated; the sizes of the uniform noise for logP-axis, tPSA-axis, and QED-axis are applied to not exceed the size of the cell sides in each axis direction.

**Latent Variable Sampling.**

As mentioned earlier, the dimension of latent variables are set to 128; 128-number of latent variables. However, since the number of tokens constituting a SMILES string is various for each molecule, the sequence length of the latent variables is applied differently each time; $\mathbb{R}^{128 \times sequence\_length}$. Here, the sequence length means the number of tokens constituting the SMILES string. The sequence length used for each molecular generation was sampled from a normal distribution. The mean and variance of the normal distribution were derived from the number of tokens constituting the SMILES strings in the MOSES training dataset. After the sequence length is determined, the values of the latent variables are sampled from the standard normal distribution.

# ■ RESULTS AND DISCUSSION

**Table 1 Comparison of the MOSES benchmarking results**

| | | | GCT (ours) | | MOSES Reference Models | | | | |
|---|---|---|---|---|---|---|---|---|---|
| | | | GCT-WarmUp | GCT-SGDR | VAE[9,30] | AAE[30,45] | Char RNN[9,30] | Latent GAN[16,30] | JTN-VAE[30,46] |
| Validity[a] | ↑ | | 0.9853 | **0.9916** | 0.9767 ±0.0012 | 0.9368± 0.0341 | 0.9748 ±0.0264 | 0.8966 ±0.0029 | 1.0±0.0 |
| Unique@1k[b] | ↑ | | **1.0** | 0.998 | **1.0±0.0** | **1.0±0.0** | **1.0±0.0** | **1.0±0.0** | **1.0±0.0** |
| Unique@10k[c] | ↑ | | 0.9981 | 0.9797 | 0.9984 ±0.0005 | 0.9973 ±0.002 | **0.9994 ±0.0003** | 0.9968 ±0.0002 | 0.9996 ±0.0003 |
| IntDiv[d] | ↑ | | 0.8531 | 0.8458 | 0.8558 ±0.0004 | 0.8557 ±0.0031 | 0.8562 ±0.0005 | **0.8565 ±0.0007** | 0.8551 ±0.0034 |
| Filters[e] | ↑ | | 0.9956 | **0.9982** | 0.9970 ±0.0002 | 0.9960 ±0.0006 | 0.9943 ±0.0034 | 0.9735 ±0.0006 | 0.9760 ±0.0016 |
| Novelty[f] | ↑ | | 0.8144 | 0.6756 | 0.6949 ±0.0069 | 0.7931 ±0.0285 | 0.8419 ±0.0509 | **0.9498 ±0.0006** | 0.9143 ±0.0058 |
| SNN[g] | ↑ | Test | 0.6179 | **0.6513** | 0.6257 ±0.0005 | 0.6081 ±0.0043 | 0.6015 ±0.0206 | 0.5371 ±0.0004 | 0.5477 ±0.0076 |
| | | TestSF | 0.5771 | **0.5990** | 0.5783 ±0.0008 | 0.5677 ±0.0045 | 0.5649 ±0.0142 | 0.5132 ±0.0002 | 0.5194 ±0.007 |
| FCD[h] | ↓ | Test | 0.4017 | 0.7980 | 0.0990 ±0.0125 | 0.5555 ±0.2033 | **0.0732 ±0.0247** | 0.2968 ±0.0087 | 0.3954 ±0.0234 |
| | | TestSF | 0.8031 | 0.9949 | 0.5670 ±0.0338 | 1.0572 ±0.2375 | **0.5204 ±0.0379** | 0.8281 ±0.0117 | 0.9382 ±0.0531 |
| Frag[i] | ↑ | Test | 0.9973 | 0.9922 | 0.9994 ±0.0001 | 0.9910 ±0.0051 | **0.9998 ±0.0002** | 0.9986 ±0.0004 | 0.9965 ±0.0003 |
| | | TestSF | 0.9952 | 0.8562 | **0.9984 ±0.0003** | 0.9905 ±0.0039 | 0.9983 ±0.0003 | 0.9972 ±0.0007 | 0.9947 ±0.0002 |
| Scaf[j] | ↑ | Test | 0.8905 | 0.8562 | **0.9386 ±0.0021** | 0.9022 ±0.0375 | 0.9242 ±0.0058 | 0.8867 ±0.0009 | 0.8964 ±0.0039 |
| | | TestSF | 0.0921 | 0.0551 | 0.0588 ±0.0095 | 0.0789 ±0.009 | **0.1101 ±0.0081** | 0.1072 ±0.0098 | 0.1009 ±0.0105 |

MOSES benchmarking was performed on 30,000 SMILES strings generated by GCT. Then, the scores obtained from GCT were compared with the scores of other generative models provided by the MOSES benchmarking platform: VAE, AAE, CharRNN, LatentGAN, and JTN-VAE. Ten indicators were used to evaluate the quality of the resulting molecules (a-j). [a]The ratio of valid SMILES strings. [b]The ratio of unique samples out of 1,000 generated molecules. [c]The ratio of unique samples out of 10,000 generated molecules. [d]The internal diversity of the generated molecules. [e]The ratio of passing through toxic substance filters. [f]The ratio of generated molecules that do not exist in the training data. [g]The average Tanimoto similarity for all generated molecules; the similarity between a generated molecule and the most similar molecule among the reference data is calculated. [h]Measurement of the distance between the generated molecules and the reference molecules using the activation of the penultimate layer of the ChemNet bioactivity prediction model. [i]The cosine similarity between the frequency in which the BRICS appear in the reference data and the frequency appearing in the generated molecules. [j]The cosine similarity between the frequency in which specific scaffolds appear in the reference data and the frequency appearing in the generated molecules. The up arrow means that a higher value is better, and the down arrow means that a lower value is better.

## How Plausible Are the Generated Molecules?

To evaluate whether the generated SMILES strings represent plausible molecular structures, analysis from two perspectives is required. The first analysis is whether the generated SMILES strings can generate valid molecular graphs, in other words, whether the generated SMILES strings satisfy both the chemical valence rule and the syntax of the SMILES language. From the benchmarking results, it was found that more than 98.5% of the generated SMILES strings are valid; GCT-WarmUp shows the validity of 98.5% and GCT-SGDR shows the validity of 99.2%. It is the highest value among language-based models (Table 1a). This ability depends on how well the generative machine can understand the geometry of molecules through SMILES strings. To determine the character (corresponding to atom, bond, or branch) followed by the given chemical string that satisfies the chemical valence rule and the grammar of the SMILES language, the geometry of the molecules (connectivity of each atom and the branches) present in the string must be understood. It seems that the attention mechanism applied to GCT helps the neural network to understand the grammar of chemical language beyond the semantic discontinuity of the SMILES language. It tends to be consistent with the results (visualized example of attention score for diproxadol) shown in ref [47].

Figure 3 shows the results for two extreme examples of how to pay attention to the characters within the SMILES string. Atom1, atom2, and atom13 in Figure 3a are located close to each other in the molecular graphs but far away from each other within the SMILES string. Although only a SMILES string was provided to GCT, it is recognized that atom1, atom2, and atom13 are related to each other (♠); Figure 3b shows that GCT-SGDR recognizes the relationship between atom2 and atom22, 23 (♦). It also recognizes atoms corresponding to a particular branch (♣) and recognizes the ring type of branch (♥). Each attention-head recognizes the molecular structure according to different viewpoints. In addition to the attention-heads visualized in **Figure 3**a and **Figure 3**b, the other attention-heads are shown in **Figure S2**. In summary, the attention mechanism applied to GCT seems to help GCT to recognize the molecular structure hidden in the one-dimensional text. This claim is consistent with the claim of ref. [47] regarding the role of attention mechanism.

The second analysis is how similar the molecular structures parsed from the valid SMILES strings are similar to real molecules. Four metrics can measure this: the SNN, FCD, Frag, and Scaf. Four metrics were evaluated on two test sets present in MOSES: Test set and TestSF set. Molecules constituting the test set have the same scaffolds of molecules constituting the train set. On the other hand, TestSF consists of molecules including scaffolds that are not included in the train set. Except for SNN (Test) and SNN (TestSF) of GCT-SGDR, and Scaf (TestSF) of GCT-WarmUp, VAE shows better scores than GCT. However, compared to the scores of other reference models, it doesn't seem that the scores of GCT are poor. SNN (Test) and SNN (TestSF) of GCT-WarmUp are higher than the score of all the other reference models except VAE. FCD (Test) of GCT-WarmUp is better than AAE and FCD (TestSF) of GCT-WarmUp is better than FCD (TestSF) of AAE, LatentGAN, and JTN-VAE. Frag (Test) and Frag (TestSF) of GCT-WarmUp is higher than Frag (Test) and Frag (TestSF) of AAE and JTN-VAE. Scaf (Test) of GCT-WarmUp is higher than LatentGAN and Scaf (TestSF) of GCT-WarmUp is higher than VAE and AAE.

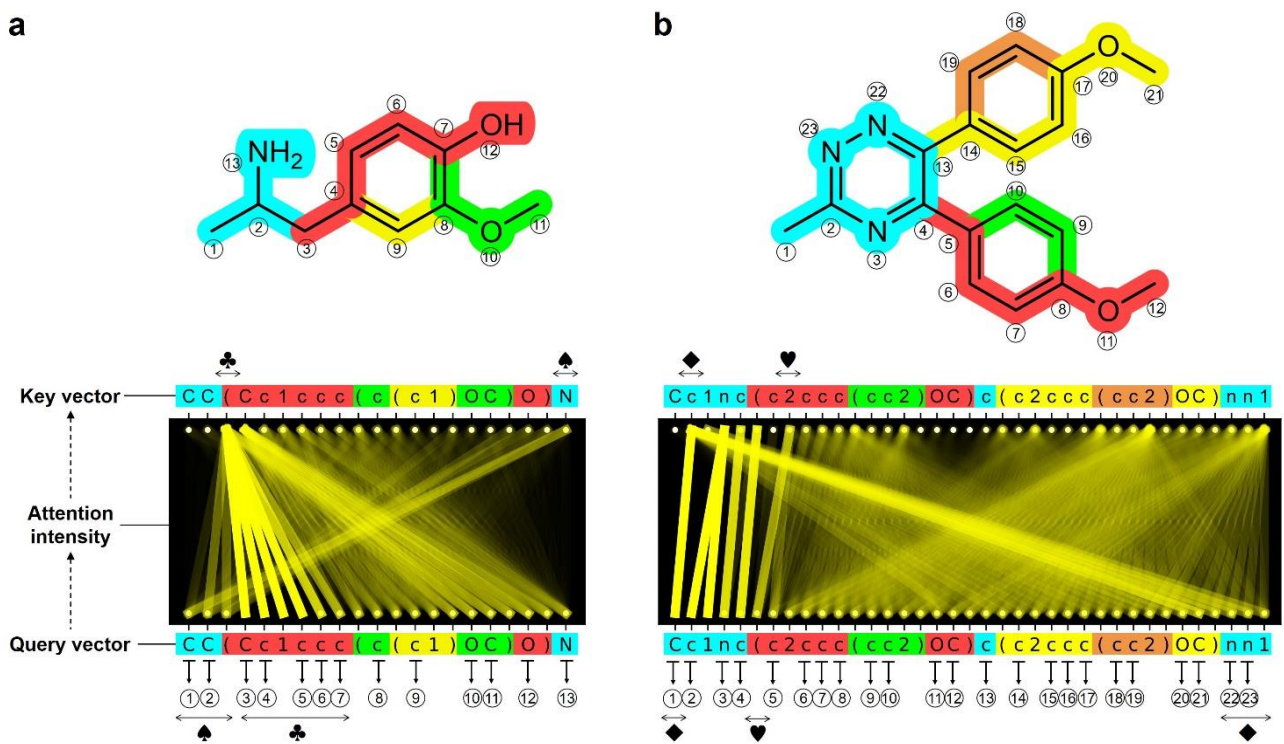

**Figure 3.** Visualized examples for self-attention scores of an attention head in encoder blocks. (a) Visualization results of the fourth head in the second encoder block for 4-(2-aminopropyl)-2-methoxyphenol. (b) Visualization results of the third head in the second encoder block for 5,6-bis(p-methoxyphenyl)-3-methyl-1,2,4-triazine. These are two extreme cases where the atom adjacent to atom2 in each molecular graph (atom13 in **a** and atom23 in **b** is far away in each SMILES string. The yellow lines in the black box at the bottom of the figure show which tokens were given a high attention score from the query vector by attention. The higher the attention score, the opaquer the yellow line. Each color in the SMILES string and the molecular graph represents each branch of the SMILES string separated with parentheses. The visualization scheme is borrowed from ref. [48].

## Do the Generated Molecules Satisfy the Target Properties?

GCT generates molecular structures that satisfy multiple target properties. **Figure 4a-c** are the results of comparing the properties of 30,000 molecules generated from GCT (calculated from RDKit) and target properties (preconditions given in GCT). Since logP and tPSA are physical properties directly related to the molecular structure, it is possible to generate a molecular structure corresponding to the target property based on an understanding of the molecular structure. However, the QED is an artificial index designed to determine the likeness to drugs quantitatively through geometric averages of eight different properties, so it is relatively difficult for the QED; this phenomenon is also found with cRNNs. The absolute mean errors between the target conditions for each property and the properties of the generated molecule are 0.177 (logP), 2.923 (tPSA), and 0.035 (QED).

The length of the generated SMILES string depends on the length of the latent code since GCT has an autoencoder (AE) structure; it is trained to reconstruct information input into the encoder. In the training phase, the length of the latent code appears equal to the length of the string input into the encoder and the length of the string output from the decoder; in fact, these are slightly different depending on whether the <sos> and <eos> tokens are used in the input and output design; however, the nature of the string is not different (see **SI**). In the inference phase, the length of the input latent code and the length of the generated SMILES string did not match perfectly and GCT does not learn the distribution of sequence lengths (**Figure 4d**). However, it seems that the length of the generated SMILES string which is related to the size of the molecule can be manipulated to some extent by adjusting the length of the latent code.

To check whether the multiple given target properties are satisfied simultaneously, the properties of generated molecules were compared to the 10 precondition sets that were sampled from the distribution of training data (**Figure 4e-g**). The conditional model, which is a skeleton of GCT, generates molecules that simultaneously satisfy multiple target properties well. Furthermore, the variational generator in GCT makes it possible to generate various molecules under the same precondition set (**Figure S3-12**).

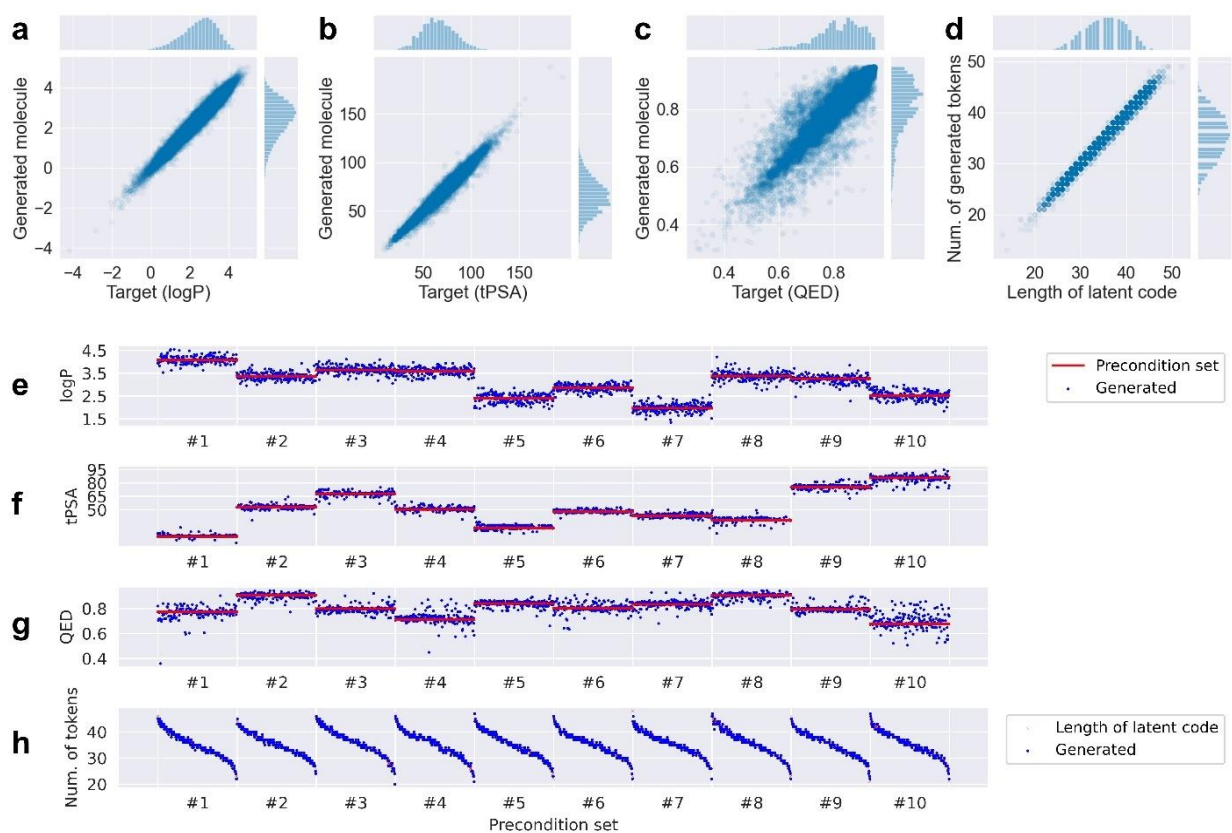

**Figure 4** Comparison of target properties (preconditions given to GCT) and properties of 30,000 generated molecules. (a)-(c), The x-axis refers to the target property, and the y-axis refers to the property of the generated molecule. (d), The x-axis refers to the length of the latent code, and the y-axis refers to the number of tokens constituting the generated SMILES string. (e)-(g), 10 precondition sets randomly sampled from the MOSES data distribution (red line) and properties of the generated molecules (blue dots). (h), The length of the latent code randomly sampled from the MOSES data distribution (red marker) and the number of tokens constituting the generated molecule (blue dots).

**Can *De Novo* Molecules Be Created?**

Whether a generative model can create de novo molecules is an important criterion that determines its applicability for material discovery. The novelty score refers to the probability of generating a new molecule that does not exist in the training data (Table 1f). Note that only a high novelty score does not guarantee that it is a good generator since odd and unrealistic molecules can increase the novelty score. Hence, the novelty score should be used in conjunction with indicators to evaluate whether the generated molecules are realistic. **Figure 5** shows a scatter plot of each model's novelty score and validity score. The dotted line is a linear regression of the reference model scores (VAE, AAE, CharRNN, GAN, JTN-VAE). Interestingly, for all reference models, it is observed that the novelty score decreases as the SNN score increases. Conversely, the higher the novelty score, the lower the SNN score. This means that it is not easy to create new molecules that are similar to real molecules. However, it can be confirmed that GCTs (GCT-WarmUp, GCT-SGDR, GCT-Exp1~3) generate new and similar molecules better than the reference models; Here, GCTs mean trained models using different hyperparameters or different learning rate schedules. The detailed conditions and scores for each GCT are summarized in Table 2.

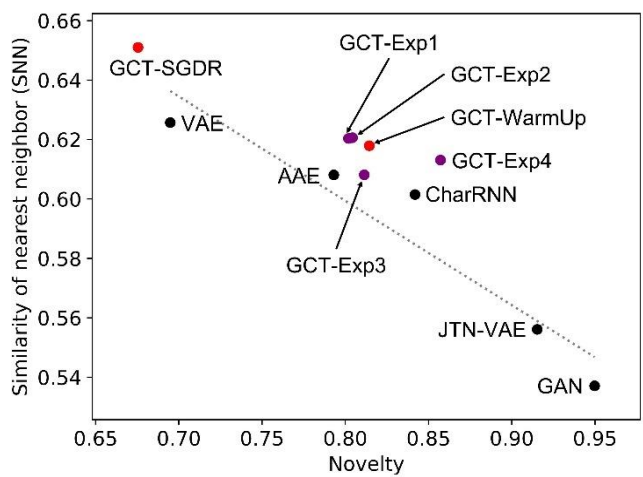

**Figure 5** Scatter plot of novelty score and similarity of a nearest neighbor score.

**Table 2** Experiments on hyperparameter tuning and learning rate scheduler type

|  | GCT-Exp1 (baseline) | GCT -WarmUp | GCT -Exp2 | GCT -Exp3 | GCT -Exp4 | GCT -SGDR |
|---|---|---|---|---|---|---|
| Training Conditions | | | | | | |
| Epochs | 25 | 25 | 25 | 25 | **50** | 25 |
| KLA weight ($k_w$) [start:step:end] | [0:0.02:0.5] | [0:0.02:0.5] | [0:0.02:0.5] | **[0:0.01:0.25]** | **[0:0.01:0.5]** | [0:0.02:0.5] |
| Learning rate (lr) | Baseline lr | **Baseline lr × 3 (= eq. 4)** | **Baseline lr × 5** | Baseline lr | Baseline lr | eq. 5 |
| Learning rate schedular | Warm-Up | Warm-Up | Warm-Up | Warm-Up | Warm-Up | **SGDR** |
| MOSES Benchmarks | | | | | | |
| Validity | 0.9757 | 0.9853 | 0.9808 | 0.9692 | 0.9813 | **0.9916** |
| Unique@1k | **1.0** | **1.0** | **1.0** | **1.0** | **1.0** | 0.998 |
| Unique@10k | 0.9977 | 0.9981 | 0.9977 | 0.9983 | **0.9986** | 0.9797 |
| IntDiv | 0.8465 | 0.8531 | 0.8531 | **0.8541** | 0.8491 | 0.8458 |
| Filters | 0.9955 | 0.9956 | 0.9950 | 0.9944 | 0.9943 | **0.9982** |
| Novelty[f] | 0.8019 | 0.8144 | 0.8043 | 0.8114 | **0.8572** | 0.6756 |
| SNN (Test) | 0.6204 | 0.6179 | 0.6206 | 0.6081 | 0.6131 | **0.6513** |
| SNN (TestSF) | 0.5784 | 0.5771 | 0.5791 | 0.5693 | 0.5744 | **0.5990** |
| FCD (Test) | 0.4181 | 0.4017 | **0.3813** | 0.3883 | 0.7180 | 0.7980 |
| FCD (TestSF) | 0.8560 | 0.8031 | 0.8411 | **0.7855** | 1.221 | 0.9949 |
| Frag (Test) | 0.9977 | 0.9973 | **0.9980** | 0.9979 | 0.9944 | 0.9922 |
| Frag (TestSF) | **0.9957** | 0.9952 | 0.9956 | 0.9960 | 0.9910 | 0.8562 |
| Scaf (Test) | **0.9021** | 0.8905 | 0.8935 | 0.8644 | 0.8893 | 0.8562 |
| Scaf (TestSF) | 0.0883 | 0.0921 | 0.1039 | **0.1049** | 0.0969 | 0.0551 |

The model using the SGDR (GCT-SGDR) shows lower novelty and higher validity than the models using the warm-up scheduler (GCT-WarmUp). A scheduler that cyclically reduces the learning rate has a loss in reducing the KL divergence term of the VAE loss function, however, it has a benefit in reducing the reconstruction error term.[49] It seems that the SGDR scheduler, a kind of cyclic annealing scheduler, makes GCT-SGCR have high validity and low novelty.

The time taken per molecule generation was 507 ms in the environment of an 8C/16T CPU and an NVIDIA GTX 1080 Ti and 440 ms with a 12C/24T and an NVIDIA Tesla T4.

**Limitations.**

Not all properties of molecules used for drug discovery were considered, and only properties of drug molecules covered by the MOSES dataset were considered; charged atoms and stereochemistry are not considered and these are limitations of this study. Furthermore, it is hard to extrapolate outside of the property window GCT was trained on since VAE, which is a model that learns the distribution of data and generates data by sampling the latent variables from the learned distribution of latent variables, cannot learn the distribution of data properly for regions where there are no data samples or for sparse regions. For this reason, the target properties are not satisfied relatively well for precondition #10 in **Figure 4**e-h; low QED area with few data samples (see **Figure 4**c).

# ■ CONCLUSIONS

In this study, a GCT architecture that embeds Transformer—a language model that has been a breakthrough in the field of NLP using an attention mechanism—into a conditional variational generator is proposed. The trained GCT can generate SMILES strings that meet the desired conditions based on a deep understanding of chemical language. It learns molecular structures and three different properties as a form of language: the logP, tPSA, and QED. Quantitative evaluations were performed by scoring molecules converted from the generated SMILES strings. In this process, the characteristics of the indicators (the SNN, FCD, Frag, and Scaf) that measure the plausibility of the molecules were analyzed, and the limitations were discussed. The performance of GCT was benchmarked by the MOSES benchmarking platform. By analyzing the results, it is demonstrated that GCT can utilize both the advantages of a language model and a conditional variational generator. The conclusions obtained are summarized as follows:

- (1) The attention mechanism in GCT helps to deeply understand the geometric structure of molecules beyond the limitations of chemical language semantic discontinuity resulting from converting a non-Hamiltonian molecular graph to a one-dimensional string by paying sparse attention to chemical formulas.

- A deep understanding of chemical language makes the generated SMILES strings (2) satisfy the syntax of SMILES language, (3) satisfy the chemical rules, and (4) are realistic.

- The conditional variational generator in GCT makes the generated molecules (5) satisfy multiple target properties simultaneously and (6) vary.

- The AE structure of GCT (7) makes the molecular size controllable.

- GCT (8) creates de novo molecules that have never been seen in the training process, and (9) creates a molecule in hundreds of milliseconds.

Well-trained GCT (GCT-WarmUp) generates valid SMILES strings with 98.5% probability. 84.1%

of the generated SMILES strings were new molecules that had never been learned, and their similarity with real molecules was 0.681 (their SNN score was 0.681). It is difficult to create a new molecule that is similar to the pattern of existing real molecules but has never been seen before. Among the compared models, GCT showed the best performance in making new molecules that are similar to real molecules. Additionally, the generated molecules satisfied multiple target properties simultaneously, and the mean absolute errors for the three different properties were 0.177 (logP), 2.923 (tPSA), and 0.035 (QED). In addition, it was confirmed that GCT can control the molecular size; the averaged difference in the number of generated SMILES tokens compared to the given length of latent code is 0.332. Molecular generation took 507 ms per molecule on a personal computer. Furthermore, conditions applicable to GCT can be configured differently as needed, and GCT can be extended to Transformer-based architectures such as BERT,[50] GPT,[51] and T5[52]. Considering the time required versus the advantages listed above, it is expected that our proposed model can contribute to accelerating the process of material discovery.

## ASSOCIATED CONTENT

**Supporting Information**

The Supporting Information is available free of charge on the ACS Publications website at DOI:

Details of the structure of GCT, input and output data structure, and generation results for the 10-precondition set (PDF)

## AUTHORS INFORMATION

**Corresponding Authors**

Jonggeol Na — Department of Chemical Engineering and Materials Science, Graduate Program in System Health Science and Engineering, Ewha Womans University, Seoul 03760, Republic of Korea; E-mail: jgna@ewha.ac.kr

Won Bo Lee — School of Chemical and Biological Engineering, Seoul National University, Gwanak-ro 1, Gwanak-gu, Seoul 08826, Republic of Korea; E-mail: wblee@snu.ac.kr

**Notes**

The authors declare no competing financial interest.

## Data and Software Availability

The datasets used for learning and evaluation of GCT were based on the MOSES benchmarking platform at https://github.com/molecularsets/moses,[30] released under the MIT license.[53] Three physical properties were calculated from RDKit. The curated data are available at https://github.com/Hyunseung-Kim/molGCT.

GCT was implemented by modifying the public code under Apache License 2.0 implementing Transformer (https://github.com/SamLynnEvans/Transformer[54]). The Python code implementing GCT is available at https://github.com/Hyunseung-Kim/molGCT.

## ACKNOWLEDGMENTS

This research was supported by the National Research Foundation of Korea (NRF) grant funded by Korean Government through the Ministry of Science and ICT (MSIT) (NRF-2018M3D1A1058633, NRF-2019R1A2C1085081, and NRF-2021R1C1C1012031).

## ■ REFERENCES

# Supporting Information
# for manuscript

**Generative chemical Transformer: Neural machine learning of molecular geometric structures from chemical language via attention**

Hyunseung Kim[†], Jonggeol Na[‡,*], Won Bo Lee[†,*]

[†]School of Chemical and Biological Engineering, Seoul National University, Gwanak-ro 1, Gwanak-gu, Seoul 08826, Republic of Korea

[‡]Department of Chemical Engineering and Materials Science, Graduate Program in System Health Science and Engineering, Ewha Womans University, Seoul 03760, Republic of Korea

**Correspondence and requests for materials should be addressed to**

J.N. (email: jgna@ewha.ac.kr) or W.B.L. (email: wblee@snu.ac.kr).

# The architecture of generative chemical Transformer

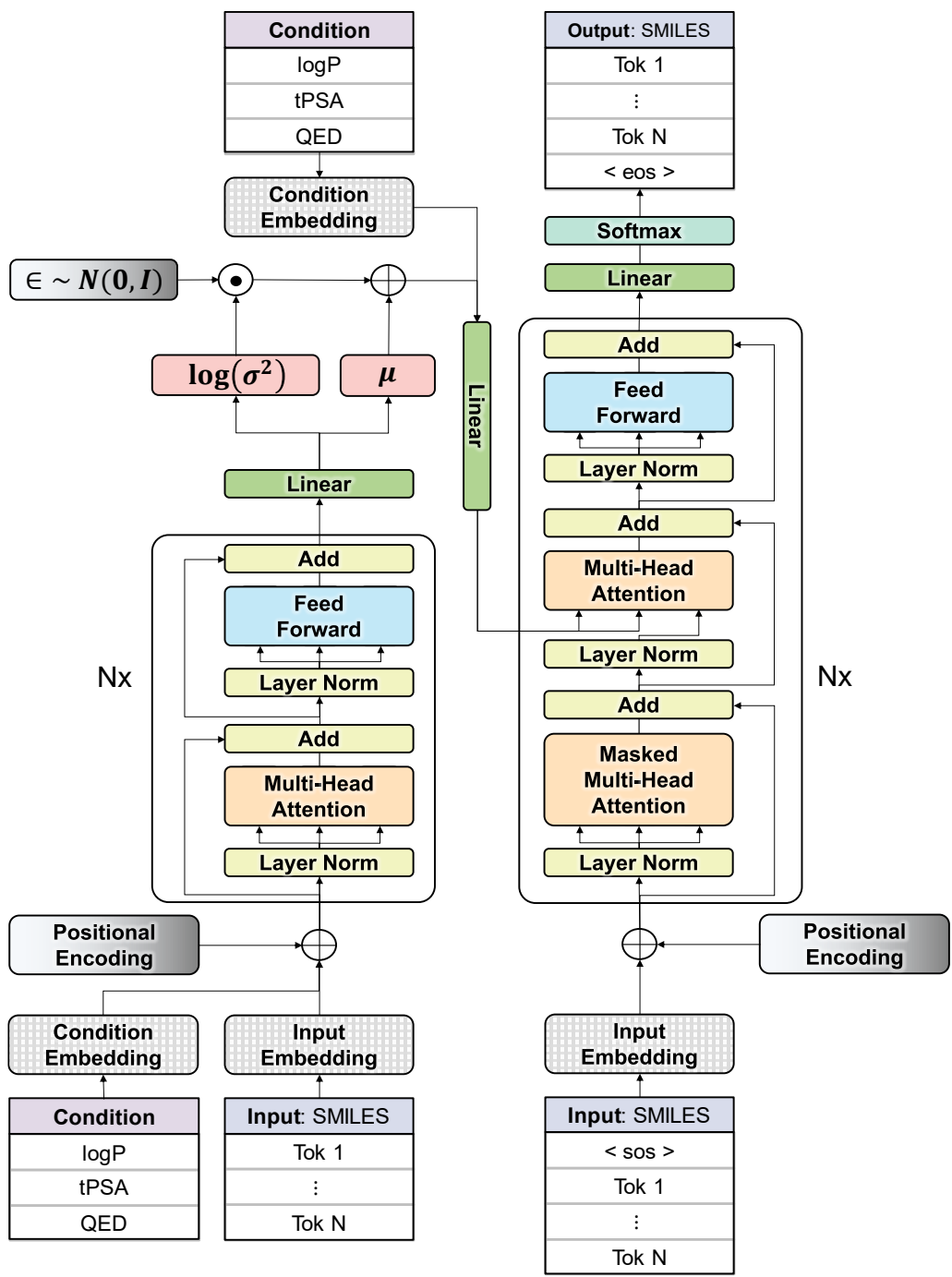

**Figure S1. The architecture of generative chemical Transformer.**

Figure S1 shows the network architecture of generative chemical Transformer (GCT). Since GCT is based on the structure of conditional variational autoencoder (cVAE)[1], it is trained in the form of reconstructing information that enters the encoder through the decoder. The encoder receives a tokenized SMILES string and three different properties: the logP, tPSA, and QED. First, the tokenized SMILES string is one-hot encoded for 26 types of tokens: <unknown>, <pad>, <C>, <c>, <O>, <o>, <N>, <n>, <F>, <S>, <s>, <Cl>, <Br>, <[nH]>, <[H]>, <->, <=>, <#>, <(>, <)>, <1>, <2>, <3>, <4>, <5>, <6>. The maximum length of the tokenized string is set to 80. The empty part is filled with <pad> token. Each token is then embedded into 512 dimensions which is the same as the model dimension. The conditions entered in the encoder are also embedded in 512 dimensions and concatenated in the embedded SMILES string; the conditions are placed in front and the SMILES string is attached to the back. The dimension entered the encoder block is then $\mathbb{R}^{(3+80)\times 512}$.

The language model applied to GCT is pre-layer normalization (Pre-LN) Transformer[2]. The encoder has six Pre-LN encoder blocks. The encoder has 8 multi-head attention, each head has 64 dimensions. Information on the properties and molecular structures leaving the Encoder is represented to the 128-dimensional conditional Gaussian latent space. Latent code sampled from latent space is concatenated with the target condition; the conditions are placed in front and the latent code is attached to the back. Then, the concatenated array expanded to 512 dimensions is entered into the encoder-decoder attention block of the decoder.

The decoder has two inputs; One is the concatenated array described above and the other input is the SMILES string generated up to the previous step. The latter is entered in the self-attention block of decoder, after one-hot encoding for 28 types of tokens: <unknown>, <sos>, <eos>, <pad>, <C>, <c>, <O>, <o>, <N>, <n>, <F>, <S>, <s>, <Cl>, <Br>, <[nH]>, <[H]>, <->, <=>, <#>, <(>, <)>, <1>, <2>, <3>, <4>, <5>, <6>. The information entered in this way passes through six pre-LN decoder blocks and exits the decoder. The array that leaves the decoder is reduced to the same dimension as the one-hot encoded tokens, and then softmax function is taken to select the next one token that will come after the string entered in the decoder's self-attention block.

The <sos> and <eos> tokens are only added to the SMILES string which is input and output to the decoder; the SMILES string input to the encoder does not have the <sos> and <eos> tokens. There are two reasons why we designed it like this. The first is that the dimension of conditions is constant, so if the SMILES string is put behind the conditions, the location of the beginning and end of the SMILES string within the concatenated input array can be recognizable. The second is the irrationality that can occur during latent code sampling. Using tokens with the <sos> and <eos> can cause problems when sampling the latent code of the <sos> and <eos> token location; the location where the <sos> appears is always constant and it cannot be randomly sampled. For this reason, the <sos> and <eos> are not used in the SMILES string entered in the encoder.

The loss function of GCT training is given as follows:

$$L(\emptyset,\theta;x_{enc},x_{<t},x_t,c) = -k_w D_{KL}(q_\emptyset(z|x_{enc},c) \parallel p(z|c)) + E_{q_\emptyset(z|x_{enc},c)}[\log g_\theta(x_t|z,x_{<t},c)] \quad (1)$$

where $\emptyset$, $\theta$, $x_{enc}$, $x_{<t}$, $z$, $x_t$, and $c$ are the parameter set of the encoder, the parameter set of the decoder, the input of the encoder, the input of the decoder, the latent variables, the target to reconstruct, and the conditions, respectively. $D_{KL}(\cdot)$ is KL divergence, and $E[\cdot]$ is the expectation of $\cdot$. $q_\emptyset$ is parameterized encoder function, $g_\emptyset$ is parameterized decoder function (generator), and $p(\cdot|c)$ is conditional Gaussian prior. $k_w$ is the weight for KL annealing. The first term in Equation (1) is a term that forces the distribution of the encoder's output to follow the conditional Gaussian prior. The second term is a term to reconstruct the target that will come after the string entered in the decoder's self-attention block using $z, c,$ and $x_{<t}$. Learning was carried out in teacher forcing manner using a mask.

## Visualization of attention heads

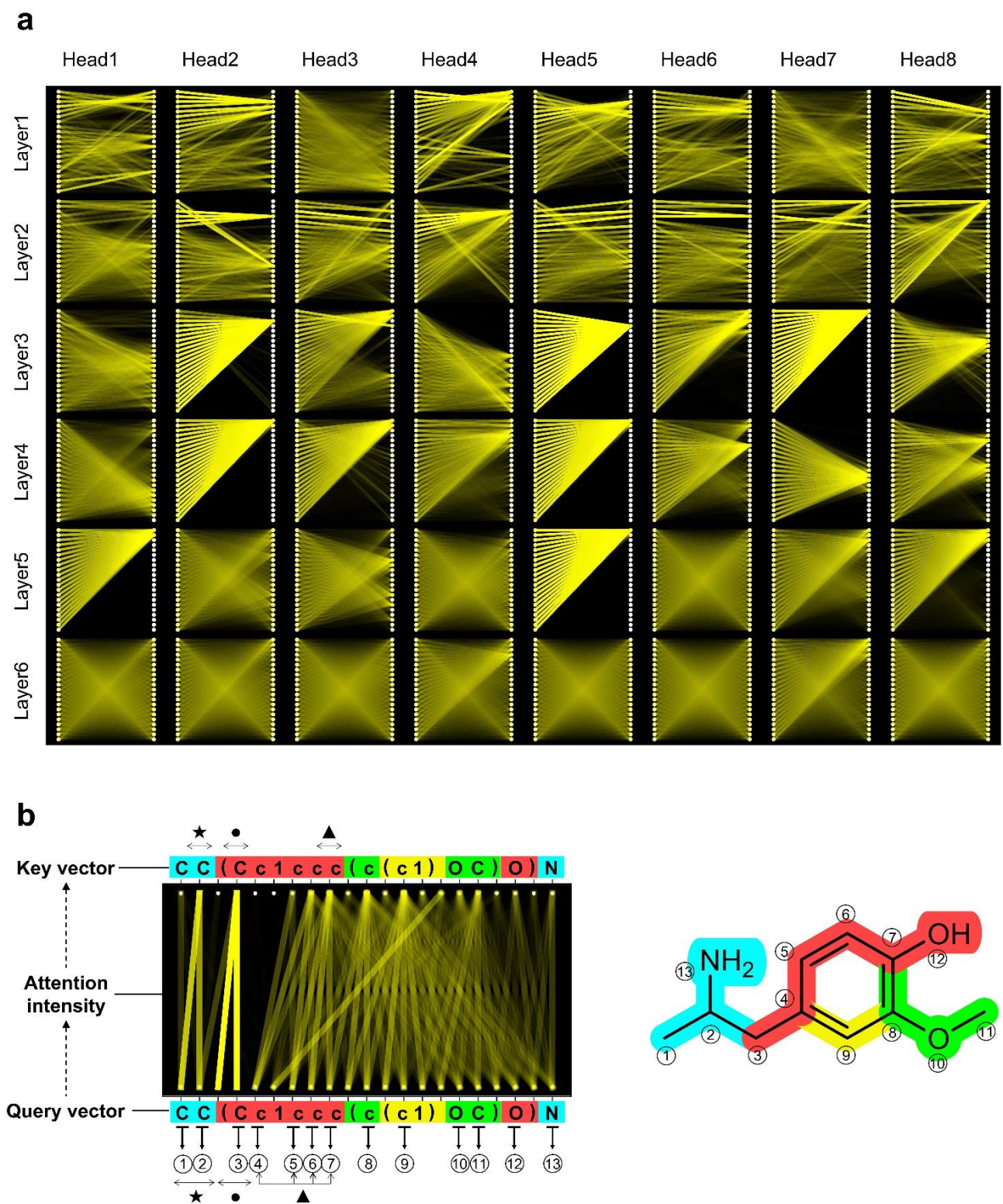

Figure S2. Visualized examples for self-attention scores of an attention head in encoder blocks for 4-(2-aminopropyl)-2-methoxyphenol. (a) Visualized attention scores for all attention heads. (b) Visualized results of the sixth head in the second encoder layer.

Figure S2a shows the visualized results of all attention heads in encoder block for 4-(2-aminopropyl)-2-methoxyphenol; note that **Figure 3**a in the main text shows the visualized results of the fourth head in the second encoder layer for 4-(2-aminopropyl)-2-methoxyphenol. Each attention-head recognizes the molecular structure according to different viewpoints. However, attention between atom 1, atom2, and atom 13, which is intuitively easy to be interpreted, is found in many attention heads. Even attention heads that don't have this connection seem to understand the molecular structure in different ways. **Figure S2**b shows the visualized results of the sixth head in the second enconder layer for 4-(2-aminopropyl)-2-methoxyphenol. It seems to recognize the connection between atom1 and atom2 (★). And it seems to recognize that atom4, 5, 6, and 7 are connected (▲). And it seems to recognize that the red branch is related to atom3 (●).

## 10-precondition set test

**Table S1 10-precondition set (Figure 4 of main text)**

| Set # | #1 | #2 | #3 | #4 | #5 | #6 | #7 | #8 | #9 | #10 |
|---|---|---|---|---|---|---|---|---|---|---|
| logP | 4.074 | 3.370 | 3.645 | 3.588 | 2.398 | 2.865 | 1.969 | 3.379 | 3.263 | 2.518 |
| tPSA | 20.26 | 53.00 | 68.00 | 50.51 | 29.72 | 47.93 | 43.19 | 38.38 | 75.36 | 85.80 |
| QED | 0.772 | 0.907 | 0.798 | 0.713 | 0.840 | 0.802 | 0.834 | 0.905 | 0.793 | 0.677 |

**Table S1** shows the details of precondition sets used in the test shown in **Figure 4** of the main text. The sets were determined by dividing 1,000 intervals between the minimum and maximum values for each property, drawing a three-dimensional histogram, and random sampling from the probability that training samples exist in each cell.

**Figures S3-12** show the randomly sampled molecules that are generated with each precondition set.

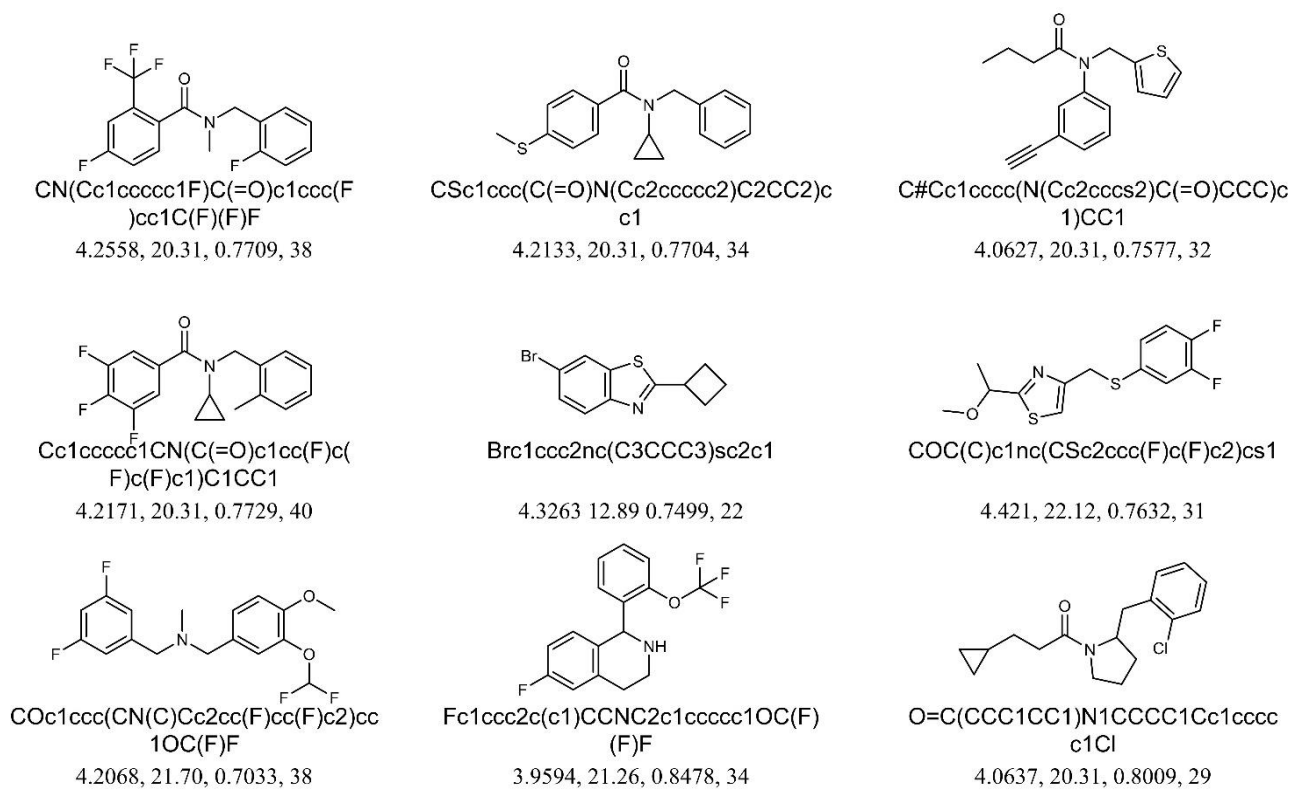

**Figure S3. Generated molecules at the given precondition set #1.** The string below the molecule image is the SMILES string. The four numbers below SMILES are actual logP, tPSA, QED, and the number of tokens, respectively.

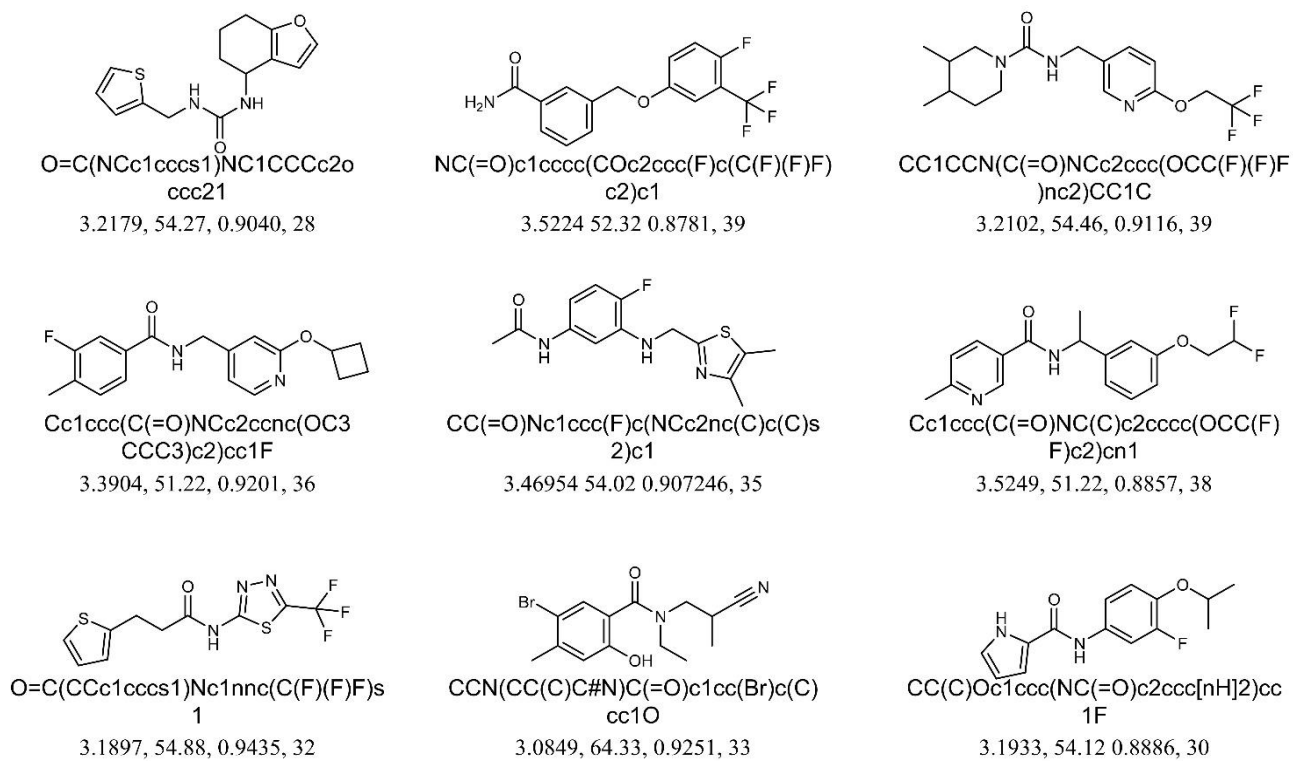

**Figure S4. Generated molecules at the given precondition set #2.**

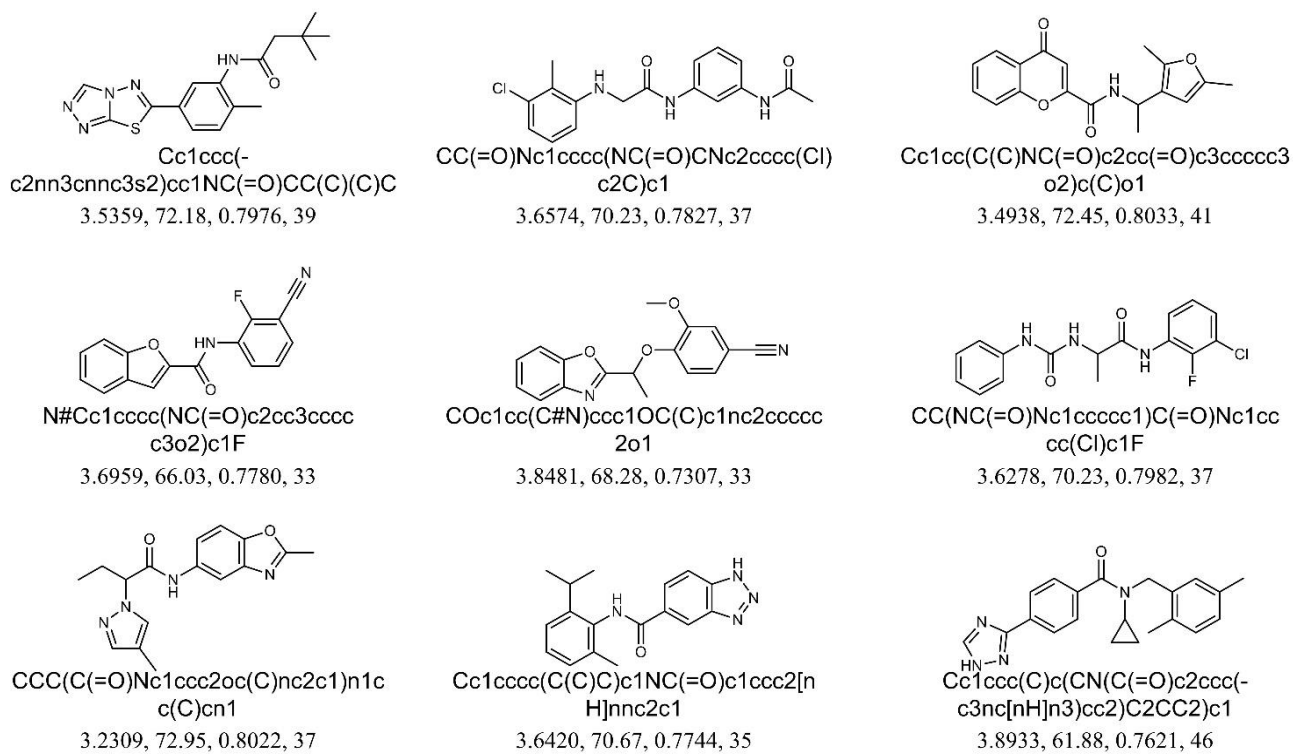

**Figure S5.** Generated molecules at the given precondition set #3.

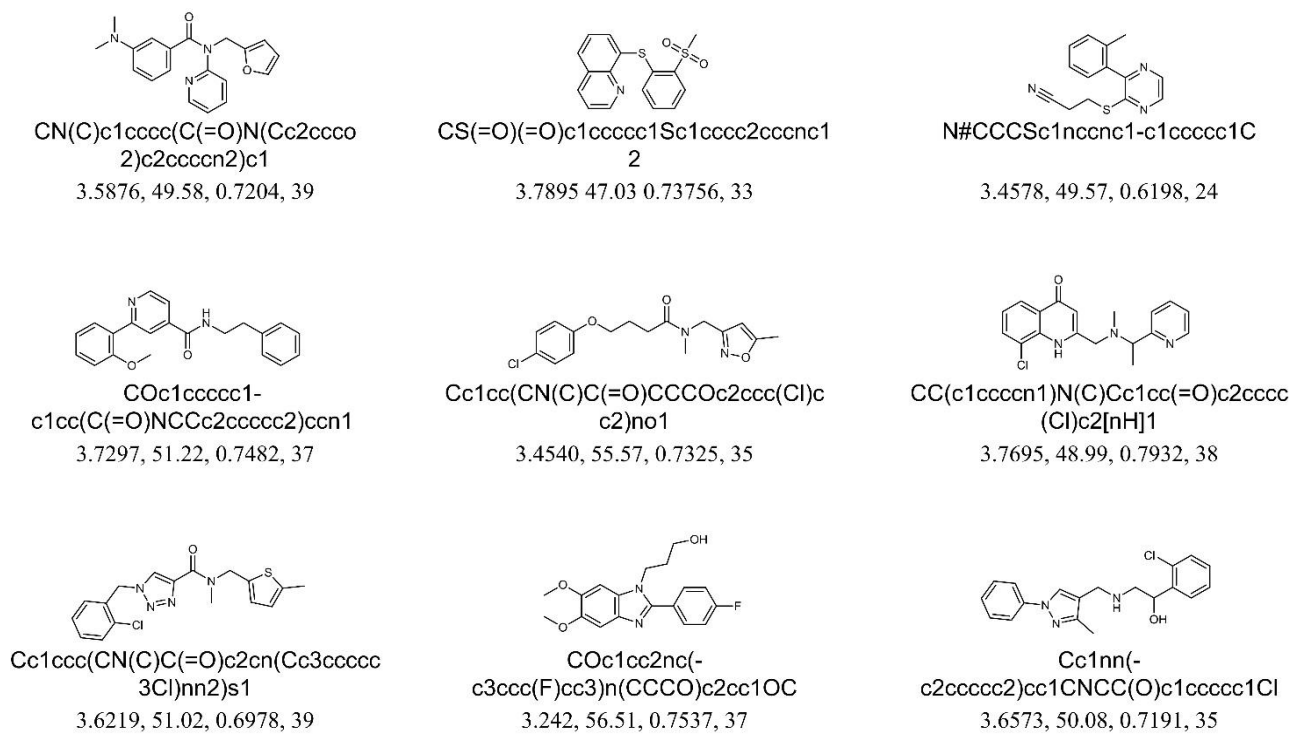

**Figure S6.** Generated molecules at the given precondition set #4.

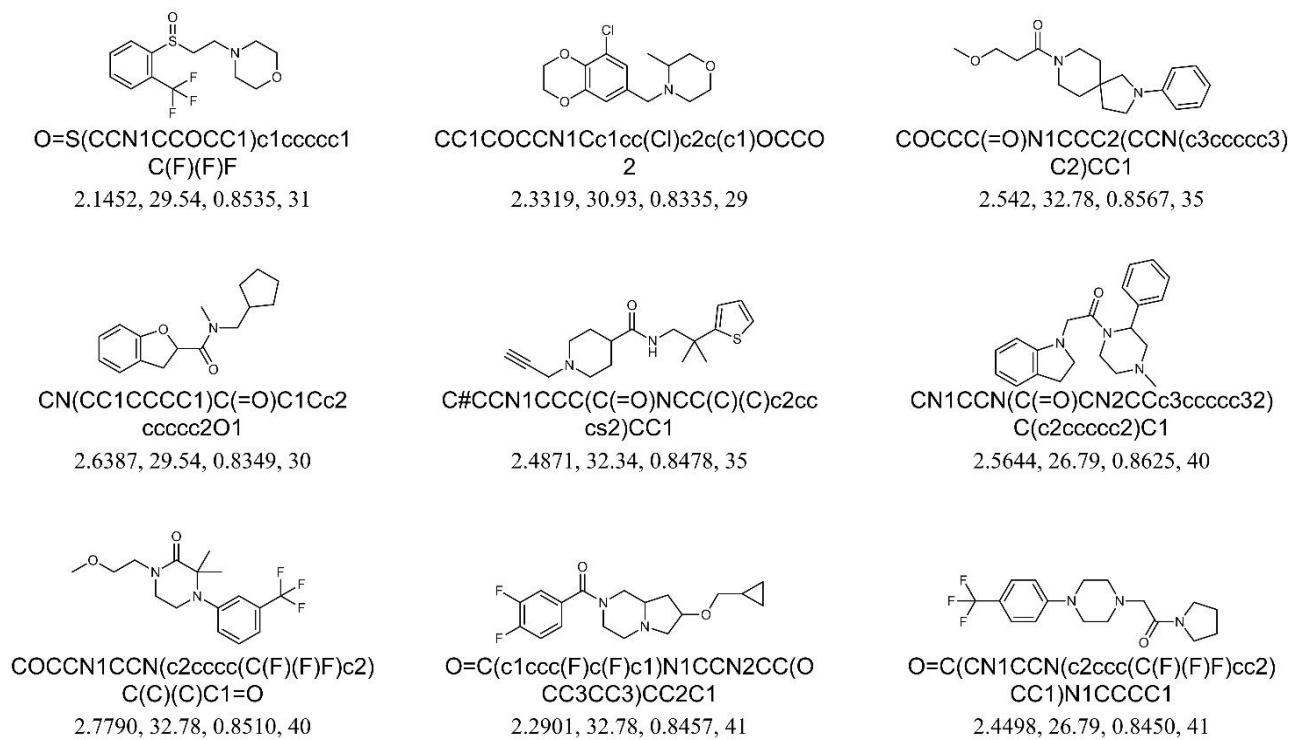

Figure S7. Generated molecules at the given precondition set #5.

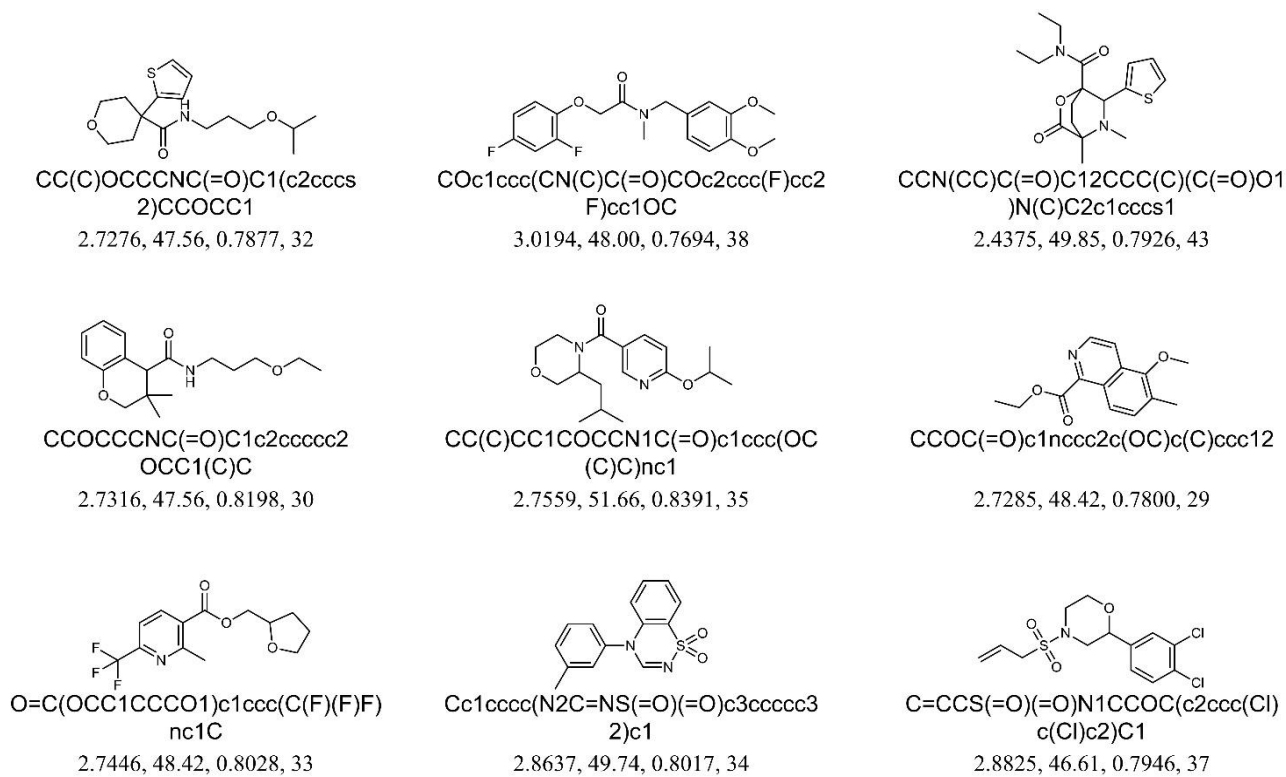

Figure S8. Generated molecules at the given precondition set #6.

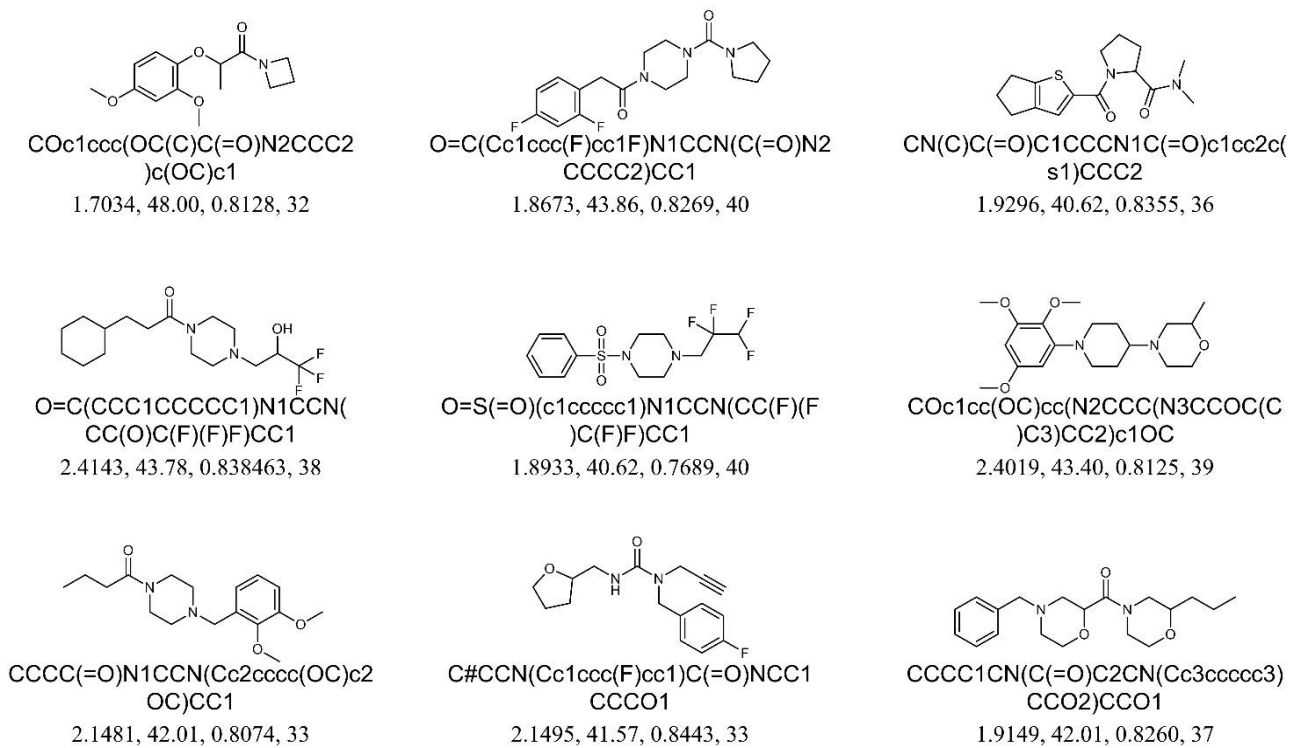

**Figure S9. Generated molecules at the given precondition set #7.**

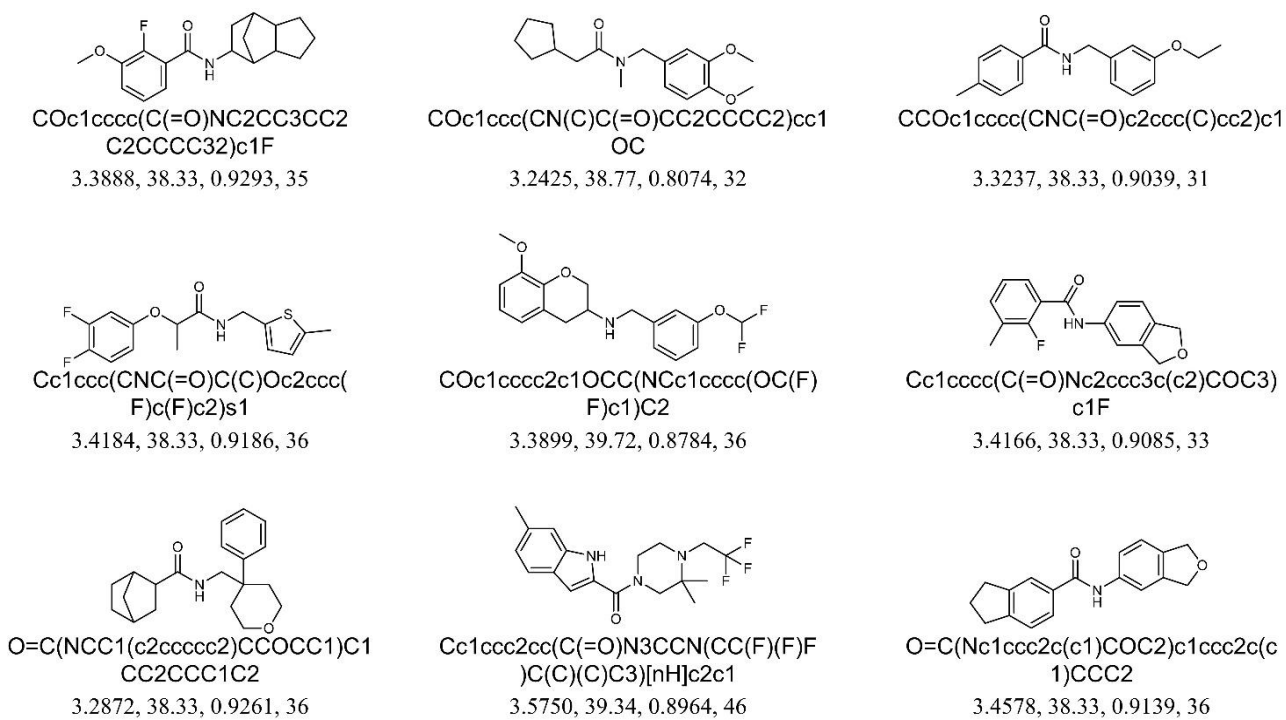

**Figure S10. Generated molecules at the given precondition set #8.**

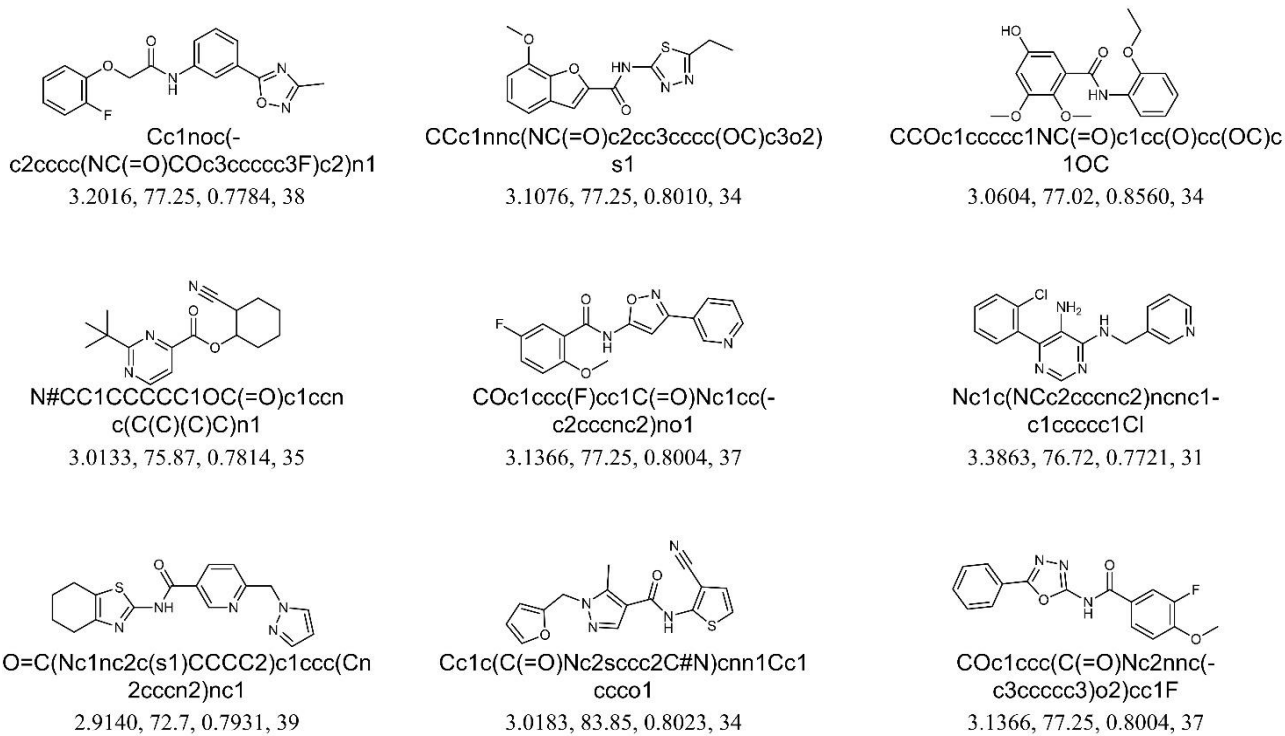

**Figure S11. Generated molecules at the given precondition set #9.**

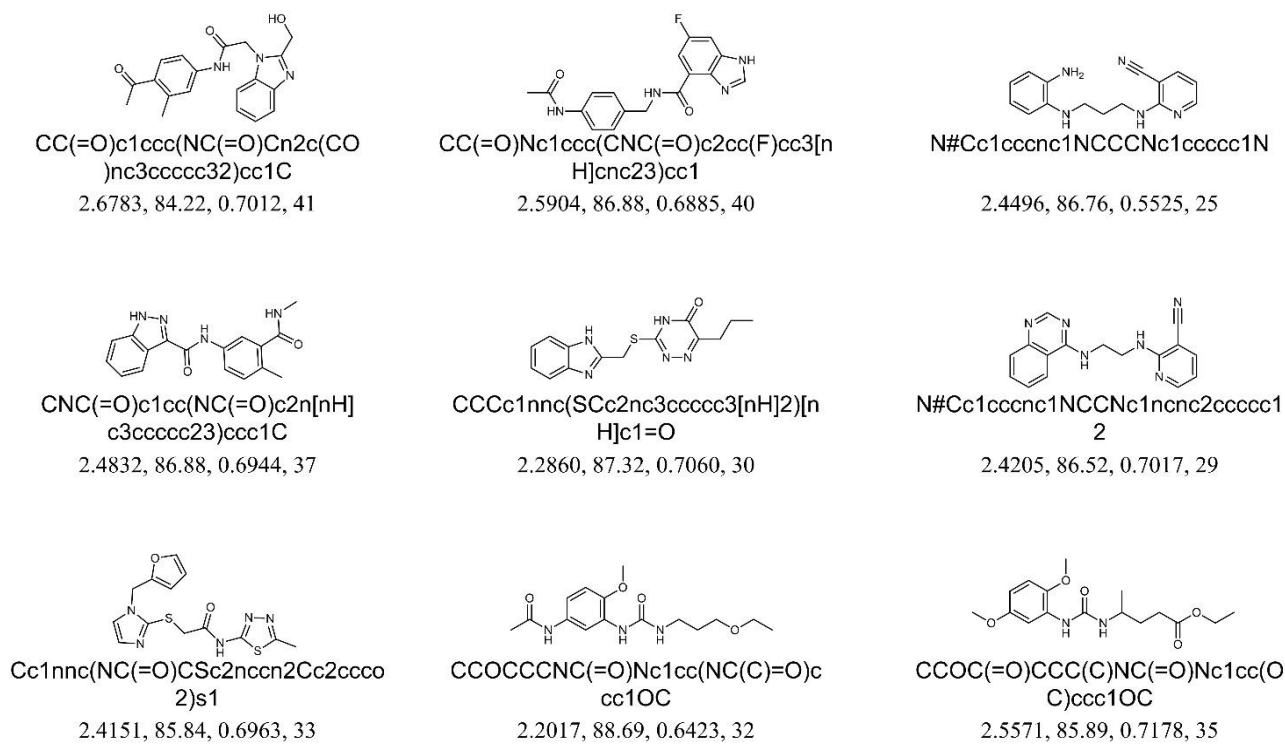

**Figure S12. Generated molecules at the given precondition set #10.**